\documentclass{article}
\usepackage{spconf,amsmath,graphicx}
\usepackage{graphicx}
\graphicspath{{img/}}
\usepackage{amsmath}
\usepackage{amssymb}
\usepackage{booktabs}
\usepackage{amsthm}
\usepackage{caption}
\usepackage{amsmath}
\usepackage[utf8]{inputenc} 
\usepackage[T1]{fontenc}    
\usepackage{url}            
\usepackage{booktabs}       
\usepackage{amsfonts}       
\usepackage{nicefrac}       
\usepackage{microtype}      
\usepackage{xcolor}         
\usepackage{subcaption}
\usepackage{caption}
\usepackage{amsmath}
\usepackage{amsthm}
\usepackage{multirow}
\usepackage{cite}
\usepackage{graphicx}

\usepackage[breaklinks,colorlinks]{hyperref}
\hypersetup{colorlinks,linkcolor={blue},citecolor={blue},urlcolor={blue}}

\usepackage[capitalize]{cleveref}

\title{
Test-Time Adversarial Detection and Robustness for Localizing Humans using Ultra Wide Band Channel Impulse Responses
}
\name{Abhiram Kolli, Muhammad Jehanzeb Mirza, Horst Possegger, Horst Bischof \thanks{This work was funded by the Austrian Research Promotion Agency (FFG) under the research project SEAMAL-Front (No.: 880598).}}
\address{Institute of Computer Graphics and Vision, Graz University of Technology, Austria.}

\begin{document}
%
\maketitle
\begin{abstract}
Keyless entry systems in cars are adopting neural networks for localizing its operators. Using test-time adversarial defences equip such systems with the ability to defend against adversarial attacks without prior training on adversarial samples. 
We propose a test-time adversarial example detector which detects the input adversarial example through quantifying the localized intermediate responses of a pre-trained neural network and confidence scores of an auxiliary softmax layer. Furthermore, in order to make the network robust, we extenuate the non relevant features by non-iterative input sample clipping. Using our approach, mean performance over 15 levels of adversarial perturbations is increased by 55.33\% for the fast gradient sign method (FGSM) and 6.3\% for both the basic iterative method (BIM) and the projected gradient method (PGD).
\end{abstract}
\section{Introduction}
\label{sec:intro}
Digital keys or keyfobs are widely used to unlock cars using UWB sensors. Typically, the channel impulse responses (CIRs) of multiple UWB sensors in and around the car are analysed by a neural network to localize these keys and verify the access permissions.
However, such systems are very vulnerable to a wide range of attacks on protocol level \cite{singh2021security, leu2022ghost,classen2022evil} and for neural network level ~\cite{szegedy2013intriguing}.
The vulnerabilities of such systems that use neural network can be mitigated by making the networks robust towards the attacks. 
Traditionally, these approaches rely on adversarial training~\cite{bai2021recent}.
Although these methods provide a certain level of robustness, they have plenty of limitations. 
For example, they require a maximum bound of the attack~\cite{moosavi2016deepfool} and need to go through an extensive adversarial training phase~\cite{moosavi2016deepfool}, which might reduce the test accuracy of the neural networks~\cite{moosavi2016deepfool, yang2020closer}. 
However, even after this tedious training process, if the specifications of the attack changes, the network again becomes vulnerable to attacks.

\begin{figure}[t!]
    \centering
    \includegraphics[width=0.45\textwidth]{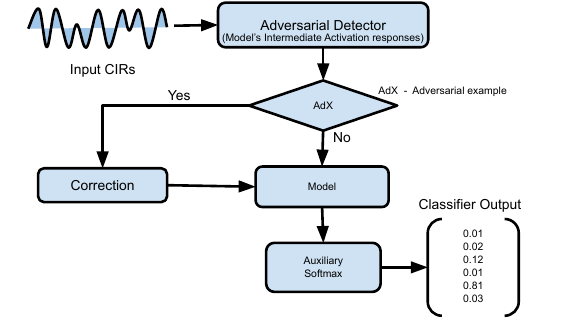}
    \caption{An overview of the adversarial example detector and robustness architecture.}
    \label{fig:archi}
\end{figure}

In contrast to adversarial training, an effective alternative is test-time robustness to adversarial attacks which either adapts the model to the incoming inputs or modify the inputs~\cite{croce2022evaluating, zhu2021efficient}. All these defensive mechanisms are iterative methods which reduce the throughput and also increase the computational resources, thus prohibiting their application in edge computing devices, which are required for the keyfob access verification in vehicles.

To address these issues, we propose an adversarial example detector which improves robustness using the model's intermediate responses. We quantify the discrepancy between these intermediate responses and use it as temperature parameter to an auxiliary softmax. This determines the classifier's confidence towards the input signal and whether an input is adversarially perturbed or not. We also propose a novel non-iterative input clipping method for test-time robustness. Since it is a non-iterative method, it improves throughput over iterative methods and is more energy efficient. Our method achieves adversarial robustness by detecting and extenuating a probable adversarial signal in the input sample through intermediate activation responses. An overview of our approach is shown in Fig.\ref{fig:archi}. Several approaches~\cite{croce2022evaluating, zhu2021efficient,wang2021fighting, chen2021towards,kang2021stable,wu2021attacking,song2017pixeldefend, nie2022diffusion, mao2021adversarial, hwang2021aid, alfarra2022combating} have studied test-time iterative defences for images but to the best of our knowledge we are the first to propose a test-time non-iterative adversarial detection and defence mechanism for localizing humans within and surrounding a car using UWB CIRs for evasion attacks.

\section{Related Work}
\label{sec:related-work}
Our work is related to methods which study conventional adversarial robustness, detection, test-time defences and more specifically to approaches which manipulate the input signal to achieve adversarial robustness at test-time. Since, there are no known direct studies on adversarial attacks and robustness on CIRs for UWB sensors based keyfob location classification, we compare our work with image classification based adversarial attacks and robustness. 

\subsection{Adversarial Robustness}\label{sec:advR}
Adversarial Robustness is traditionally achieved by adding an adversarial regularization term during training~\cite{goodfellow2014explaining}, which is broadly termed adversarial training. 
Madry et al.~\cite{madry2017towards} use a natural saddle point (min-max) formulation during optimization to achieve robustness. Kannan et al.~\cite{kannan2018adversarial} proposed adversarial logit pairing during training for robustness. 
Gowal et al.~\cite{gowal2020uncovering} study different design choices for training deep neural networks to achieve robustness.
Similarly, other works, e.g.~\cite{shafahi2019adversarial} achieve robustness by designing a sophisticated method for updating model parameters and the sample through self-supervision~\cite{naseer2020self} and by generative inversion~\cite{zhang2020secret}.

\subsection{Test-Time Defences}\label{sec:ttD}
Test-Time Defences for adversarial robustness counter attacks from a different perspective -- adapt to adversarial perturbations `as and when' they are encountered.
The goal of these methods is to adapt to an attack which a neural network might become exposed to at test-time.
These methods can be divided into two categories. 
One category updates the model weights at test-time to achieve robustness. 
For example, DENT~\cite{wang2021fighting} minimizes prediction entropy as adversarial examples are encountered during test-time. 
Chen et al.~\cite{chen2021towards} modify the activation responses for robustness against attackers. 
Kang et al.~\cite{kang2021stable} design a neural ordinary differential equation and show that they are more robust than ordinary DNNs.

The second category which is more related to our work is achieving test-time defence by \textit{purifying inputs}. 
For example, Wu et al.~\cite{wu2021attacking} achieve input sample purification by attacking all classes simultaneously in such a manner that the misplaced predictions can be flipped into correct ones.
Some other approaches, e.g.~\cite{song2017pixeldefend, nie2022diffusion} use generative diffusion models to remove adversarial perturbations.
Mao et al.~\cite{mao2021adversarial} propose to achieve robustness by reversing the adversarial examples by finding the minimum reverse vector through gradient back propagation. 
Hwang et al.~\cite{hwang2021aid} add an auxiliary network for purifying inputs.  
Alfarra et al.~\cite{alfarra2022combating} combat adversaries by adding an additional layer which reverses the direction of the attack. Zhu et al. ~\cite{zhu2021efficient} proposed to add saliency maps to the inputs that extenuate the insignificant features. 
\begin{figure}[t]
    \centering
    \includegraphics[width=0.35\textwidth]{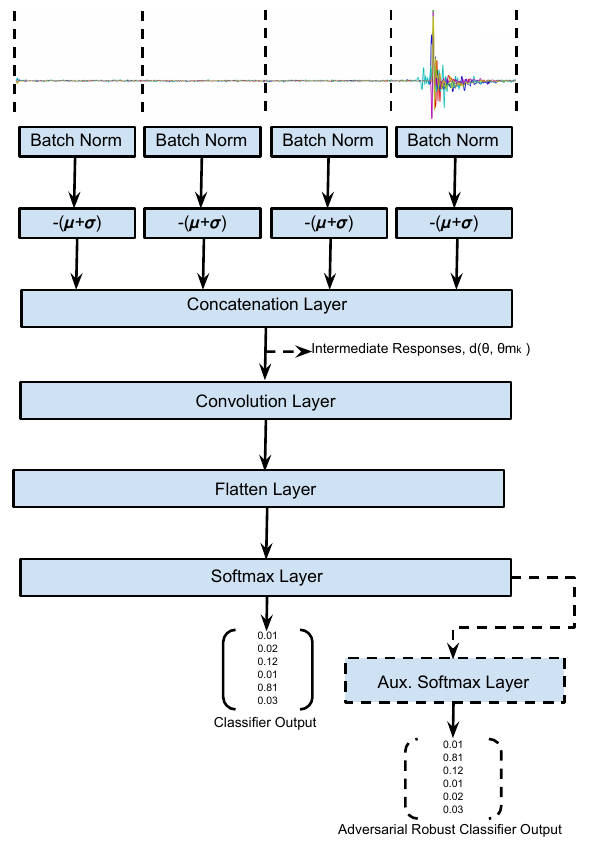}
    \caption{An overview of our network architecture for adversarial example detector and robustness.}
    \label{fig:model}
\end{figure}

Test-time robustness approaches ~\cite{mao2021adversarial, alfarra2022combating, zhu2021efficient, hwang2021aid} are closely related to our work. However, we differ to them in the following fundamental ways: First, in contrast to~\cite{mao2021adversarial, alfarra2022combating} we do not search for the bound of attack and then try to mitigate it.
Second, contrary to~\cite{zhu2021efficient}, our method does not rely on approaches such as generating saliency maps iteratively to enlarge the distance between original data and the decision boundary. Third, they  update pre-trained model's batch norm statistics by forward passing the test samples and use gradients to determine the direction to the original class. Our method obtains the direction by the sign of intermediate responses from the network.   
Fourth, Zhu et al.~\cite{zhu2021efficient} proposed a method which takes several steps to predict the adversarial distortion and the direction of the gradients. This might lead to either vanishing or exploding gradients. Since our method is non-iterative, this scenario does not arise. Fifth, multi-step approaches are not desirable in resource constrained embedded devices where UWB pulses arrive with the repetitive frequency between 1 to 50 MHz~\cite{staderini2001everything}. 
Sixth, unlike in \cite{hwang2021aid}, though we also purify the inputs we do not use any auxiliary neural network. We only use an auxiliary softmax layer which does not require any training.

\begin{figure*}
\begin{minipage}{0.48\linewidth}
  \centering
  \centerline{\includegraphics[scale=0.38,trim = 0 9 0 13, clip]{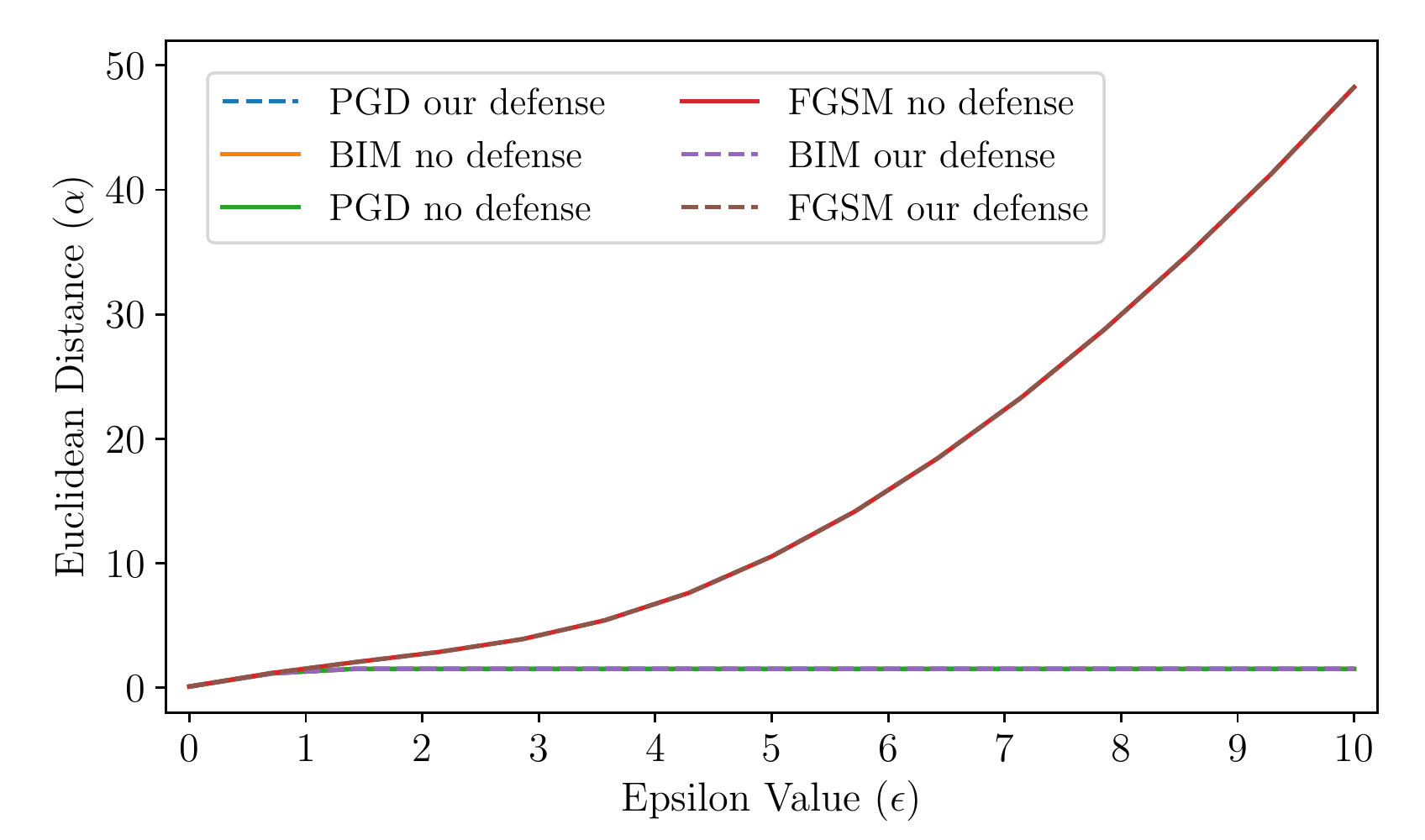}}
  \caption{Euclidean distance between the intermediate responses of the network with respect to adversarial perturbations. With the increase in perturbation higher is better.}
  \label{fig:EUDist}
\end{minipage}
\hfill
\begin{minipage}{0.48\linewidth}
  \centering
  \centerline{\includegraphics[scale=0.38,trim = 0 9 0 12, clip]{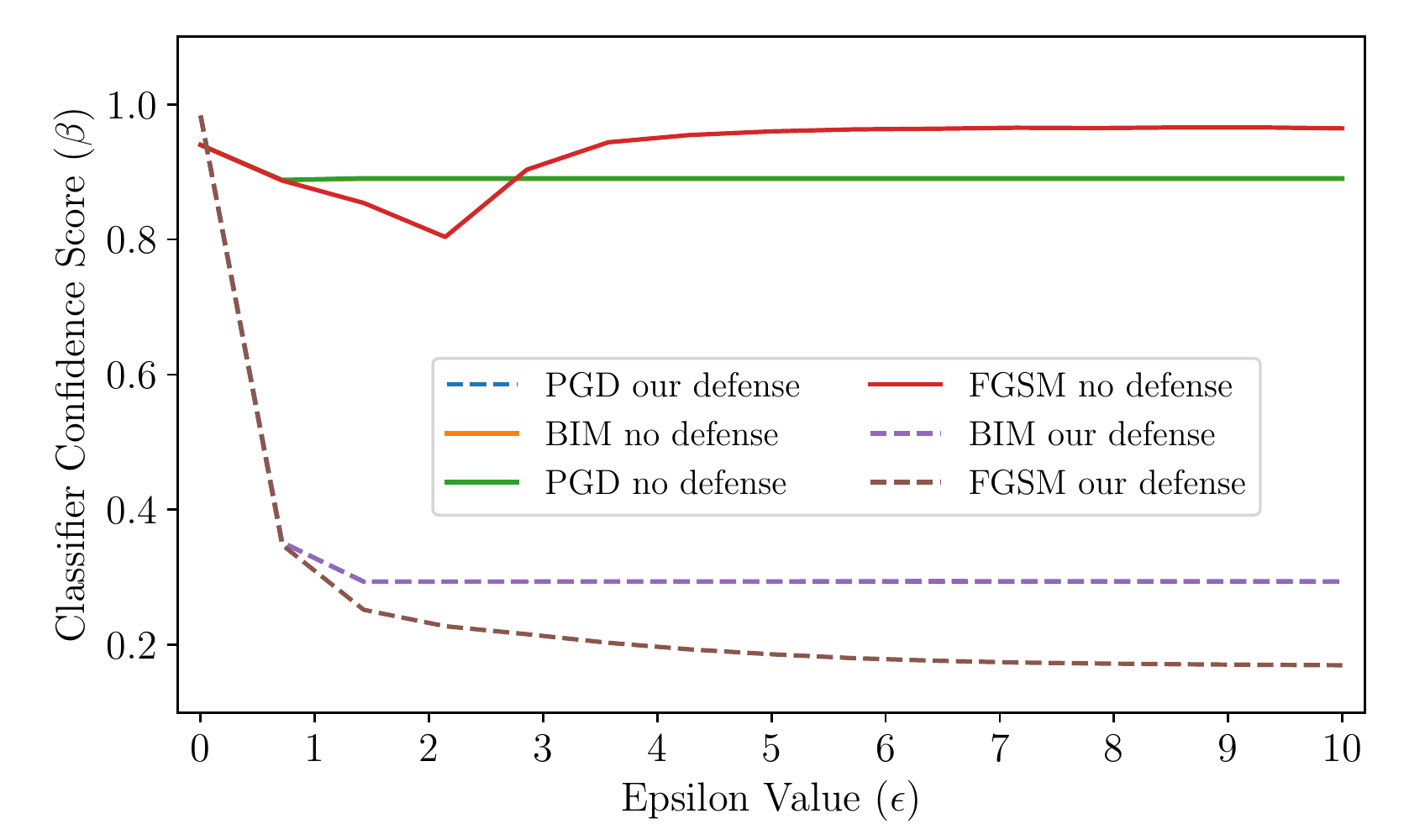}}
  \caption{Confidence scores of the classifier with respect to adversarial perturbations. With the increase in perturbation lower is better.}
  \label{fig:ConfScore}
\end{minipage}
\end{figure*}
\section{Methodology}
\label{sec:pagestyle}
In this section, we detail the architecture of our pre-trained network (Fig.~\ref{fig:model}), the mechanism of adversarial detection and adversarial robustness.

\subsection{Architecture}\label{sec:arch}


Our signal classification model (Fig.~\ref{fig:model}) consists of an input layer which receives the input UWB CIRs of dimension B~$\times$~N~$\times$~S, where B is the batch size, N is the number of sensors, S is the CIR for each sensor, 1024 in this case. This is followed by 4 parallel blocks (k = 0, 1, 2, 3) where the input to each block is a B~$\times$~N~$\times$~256 patch of non-overlapping input tensor. Each block consists of a batch normalization layer followed by the mean ($\mu$) and variance ($\sigma$) subtraction layers along the S dimension of the input (Fig.~\ref{fig:model}). 
The output of these modules are concatenated and fed to a single convolution layer consisting of a single filter of size~3~$\times$~3~with L2 regularization.
This is followed by a flattened layer and a softmax layer. The intermediate responses of the network (dashed arrows in Fig. \ref{fig:model}) are used for both adversarial detection and robustness. An auxiliary softmax layer (dashed box in Fig. \ref{fig:model}) uses the intermediate responses to quantify our classifier's confidence and outputs the robust classification results. 

\subsection{Adversarial detection}
\label{sec:advdet}

During test-time our classifier detects if an incoming sample is an adversarial example or not.  We do this by taking the Euclidean distance between the pre-trained model's intermediate responses after the concatenation layer (Fig. \ref{fig:model}).
\begin{equation}\label{eq:dist}
d(\theta, \theta_{m_{k}}) = \lVert f(x,\theta)- f(x,\theta_{m_{k}=0}) \rVert,
\end{equation}
where, $f(\cdot)$ is the output of the pre-trained model's intermediate response after the concatenation layer, $x$ is the single input sample, $\theta$ represents the weights of the intermediate response, $m_{k}$ is the momentum of the batch normalization layer, $k \in [0, \dots, 3]$
is the block number of batch normalization layer as illustrated in Fig \ref{fig:model}. The distance between clean samples is expected to be near 0 as the network is pre-trained with clean samples and increases for adversarial samples. This distance is also the temperature parameter for the auxiliary softmax layer which computes

\begin{equation}\label{eq:temp}
    \hat{y}(x_{i}) = \frac{\exp(\frac{y_{i}}{d(\theta, \theta_{m_{k}})+\zeta})}{\sum_{j}^{N} \exp(\frac{y_{i}}{d(\theta, \theta_{m_{k}})+\zeta})}
\end{equation}

where, $ \hat{y}(x_{i})$ is the output of the auxiliary softmax for class \textit{i}, ${y_{i}}$ is output of networks softmax layer for class \textit{i}, N is the total number of classes, $\zeta$ is for numerical stability (not used in these experiments). We take the Euclidean distance as the temperature parameter because for an incoming clean test sample the distance is smaller so the confidence of prediction remains mostly unchanged. When the input is an adversarial example, this distance increases which decreases the confidence of the model's prediction.

An input signal is considered as adversarial if

\begin{equation}\label{eq:advDet}
    \hat{x}=
    \left\{\begin{matrix}
        \text{clean} \phantom{0000000000000}& \quad\text{if}\,(\alpha < 0.2)\,\land \,(\beta> 0.98)\\
        \text{adversarial} \phantom{00000000} & \text{otherwise}
    \end{matrix}\right.
\end{equation}

    where, $\alpha$ is Euclidean distance from Eq.~\eqref{eq:dist} and $\beta$ is the confidence score of the auxiliary softmax from Eq.~\eqref{eq:temp}. The threshold values of 0.2 and 0.98 are y-axis values for clean samples as illustrated in the Fig. \ref{fig:EUDist} and Fig. \ref{fig:ConfScore} respectively. 
\subsection{Adversarial robustness}
\label{ssec:advrob}


After passing through the adversarial detection block, we clip those input samples which are flagged as adversarial sample by the detector.


\begin{equation}
\label{eq:samplselct}
\hat{x}=\left\{\begin{matrix} x*(\sigma(\gamma)-\operatorname{sgn}(\gamma)) & if\,(\alpha > 0.2 )\,\lor \,(\beta < 0.98)&&\\
x \phantom{000000000000000} & \text{otherwise} \phantom{0000000000000}&&\end{matrix}\right.
\end{equation}

where, \textit{x} is the incoming sample, $\sigma$ is the sigmoid activation function. $\gamma = f(x,\theta)- f(x,\theta_{m_{k}=0})$ is the difference of intermediate responses of the module stated in Section $\ref{sec:advdet}$. 

\begin{table*}
\setlength\tabcolsep{4.0pt}
\centering
\begin{tabular}{l|ccccccccccccccc}
\toprule
    Distortion level              & L1 & L2 & L3 & L4 & L5  & L6 & L7 & L8 & L9 & L10 & L11 & L12  & L13 & L14 &L15\\
    \midrule
    Source (FGSM)              & 0.65 & 0.65 & 0.66 & 0.66 & 0.66 & 0.65 & 0.64 & 0.63 & 0.62 & 0.62 & 0.62 & 0.63 & 0.63 & 0.64  & 0.65  \\
    Source (BIM)              & 0.78 & 0.77 & 0.78 & 0.78 & 0.78 & 0.79 & 0.79 & 0.79 & 0.79 & 0.79 & 0.79 & 0.79 & 0.79 & 0.79 & 0.79  \\
    Source (PGD)              & 0.78 & 0.77 & 0.78 & 0.78 & 0.78 & 0.79 & 0.79 & 0.79 & 0.79 & 0.79 & 0.79 & 0.79 & 0.79 & 0.79 & 0.79   \\
     \midrule
    Our (FGSM)              & 0.98 & 0.99 & 0.99 & 0.99 & 0.99 & 0.99 & 0.99 & 0.99 & 0.99 & 0.99 & 0.99 & 0.99 & 0.99 & 0.99 & 0.99\\ 
    Our (BIM)              & 0.98 & 0.99 & 0.99 & 0.99 & 0.99 & 0.99 & 0.99 & 0.99 & 0.99  & 0.99 & 0.99 & 0.99 & 0.99 & 0.99 & 0.99\\
    Our (PGD)              & 0.98 & 0.99 & 0.99 & 0.99 & 0.99 & 0.99 & 0.99 & 0.99 & 0.99 & 0.99 & 0.99 & 0.99 & 0.99 & 0.99 & 0.99\\
    \bottomrule

\end{tabular}
\caption{Comparison of adversarial example detection of a pre-trained neural network without and with our defensive mechanism. Here, L1 is without any distortion and L15 is the highest distortion level. 'Source' is the model without our defensive mechanism.}
\label{tab:tbl1}
\end{table*}

\begin{table*}
\setlength\tabcolsep{4.0pt}
\centering
\begin{tabular}{l|ccccccccccccccc}
\toprule
    Distortion level              & L1 & L2 & L3 & L4 & L5  & L6 & L7 & L8 & L9 & L10 & L11 & L12  & L13 & L14 &L15\\
    \midrule
    Source (FGSM)              & 0.93  & 0.80 & 0.67  & 0.28  & 0.09 & 0.06 & 0.06  & 0.06  & 0.06
     & 0.06  & 0.06 & 0.06  & 0.07  & 0.07 & 0.07  \\
    Source (BIM)              & 0.93 & 0.80    & 0.75  & 0.75  & 0.75  & 0.75  & 0.75  & 0.75  & 0.75  & 0.75
     & 0.75  & 0.75  & 0.75  & 0.75  & 0.75   \\
    Source (PGD)              &  0.93 & 0.80    & 0.75  & 0.75  & 0.75  & 0.75  & 0.75  & 0.75  & 0.75  & 0.75
     & 0.75  & 0.75  & 0.75  & 0.75  & 0.75   \\
     \midrule
    Our (FGSM)              & 0.92   & 0.83  & 0.77  & 0.66 & 0.56 & 0.57  & 0.61  & 0.70 & 0.77 & 0.83
     & 0.88  & 0.90  & 0.90 & 0.90 & 0.90 \\ 
    Our (BIM)              & 0.92   & 0.84  & 0.82 & 0.82  & 0.82  & 0.82  & 0.82 & 0.82  & 0.82  & 0.82
     & 0.82 & 0.82  & 0.82 & 0.82 & 0.82\\
    Our (PGD)              & 0.92   & 0.83 & 0.82 & 0.82 & 0.82 & 0.82 & 0.82 & 0.82 & 0.82 & 0.82 & 0.82 & 0.82 & 0.82 & 0.82 & 0.82\\
    \bottomrule

\end{tabular}
\caption{Comparison of adversarial robustness of a pre-trained neural network without and with our defensive mechanism. Here, L1 is without any distortion and L15 is the highest distortion level. 'Source' is the model without our defensive mechanism.}
\label{tab:tbl2}
\end{table*}

\section{RESULTS}
\label{sec:resu}

We evaluate our approach on a custom dataset, consisting of 461 samples where each sample consists of N=6 CIRs representing 6 sensor outputs. Among these 6 sensors, 4 sensors are deployed surrounding the car and 2 are placed within the car. These 6 CIRs together determine the presence of a keyfob into namely \emph{back}, \emph{back seat}, \emph{driver seat}, \emph{front}, \emph{left} and \emph{right}. For all the experiments in this paper, we used randomly shuffled 231 samples to pre-train the network and the remaining 230 samples are subjected to adversarial perturbations. We modeled three types of adversarial attacks namely fast gradient sign method, basic iterative methods and projected gradient descent. These attacks are modeled as mentioned in \cite{ren2020adversarial}. A total of 15 levels per each type of attack are used where level1 ($\epsilon$ = 0) is clean sample and level15 is highest adversarial perturbation ($\epsilon$ = 10). The intermediate values of $\epsilon$ used are linearly spaced between 0 through 10.

\subsection{Adversarial Detection}\label{sec:resDet}
Table~\ref{tab:tbl1} shows the performance of our detector for all classes various adversarial distortions. Performance of source model (without defence mechanism) is around 65\% for FGSM attack. With our approach the network is able to detect with an accuracy of 98$\%$ or above for FGSM attack. The performance of source model is around 79\% for both BIM and PGD attacks. Our mechanism is consistently 98\% and above with all types of attacks as shown in Table~\ref{tab:tbl1}.   

\subsection{Adversarial Robustness}\label{sec:resRob}
The classifier's adversarial robustness (over all classes) performance is shown in Table \ref{tab:tbl2}. Performance of the source model drops immediately even with small adversarial perturbations for FGSM attack. However, our defensive mechanism performs better. Though the accuracy of our model drops to 56\% at level5, it increases and stabilizes at 90\% for higher levels of adversarial perturbations for the FGSM attack. This phenomenon can be attributed to the usage of the sigmoid activation function in our input clipping function. FGSM attack has a mean performance of 78\% over 15 levels with our defence mechanism whereas it is just 22.6\% for source model. This is a 55\% improvement over the source model. The performance of the network remains consistently higher for BIM and PGD attacks at 6\% higher than the source model.

\section{Conclusion}
\label{sec:conc}
We show that adversarial examples can be detected with high accuracy even at low adversarial perturbations. With this performance level of adversarial detection, our system can also be used to discard the samples without classifying them. We also achieve high adversarial robustness with non iterative clipping function even for higher $\epsilon$ values. Since our approach is a non iterative method, it provides better throughput over iterative methods and better energy efficiency. Due to its very low memory foot print (162.1 KB, tflite model without quantization), our model with defensive mechanism can be deployed on an edge computing device that can be deployed on a vehicle.




\bibliographystyle{IEEEbib}
\bibliography{strings,refs}

\end{document}